\RequirePackage{fix-cm}
\documentclass[smallcondensed]{svjour3}     % onecolumn (ditto)
\smartqed  % flush right qed marks, e.g. at end of proof
\usepackage{graphicx}
\usepackage[vlined,linesnumbered,ruled]{algorithm2e}
\usepackage[cmex10]{amsmath}
\usepackage{hyperref,url}
\DeclareMathOperator*{\argminA}{arg\,min}

\DeclareMathOperator{\sgn}{sgn}

\journalname{Neural Processing Letters}
\begin{document}

\title{A Probabilistic Optimum-Path Forest Classifier for Binary Classification Problems%\thanks{Grants or other notes
}


\author{Silas E. N. Fernandes \and
		Danillo R. Pereira \and
		Caio C. O. Ramos \and
		Andr\'e N. Souza \and
        Jo\~ao P. Papa
}


\institute{Silas E. N. Fernandes \at
              Department of Computing, Federal University of S\~ao Carlos, S\~ao Carlos, Brazil\\
              %\email{silas.fernandes@dc.ufscar.br}%  \\
           \and
           Danillo R. Pereira \at
           University of Western S\~ao Paulo, Presidente Prudente, Brazil\\
           %\email{danilopereira@unoeste.br}
           \and
           Caio C. O. Ramos \at
           Department of Electrical Engineering, S\~ao Paulo State University, Bauru, Brazil\\
           \and
           Andr\'e N. Souza \at
           Department of Electrical Engineering, S\~ao Paulo State University, Bauru, Brazil\\
           \and
           Jo\~ao P. Papa \at
           Department of Computing, S\~ao Paulo State University, Bauru, Brazil\\
      	   \email{papa@fc.unesp.br}
}

\date{Received: date / Accepted: date}

\maketitle

\begin{abstract}
Probabilistic-driven classification techniques extend the role of traditional approaches that output labels (usually integer numbers) only. Such techniques are more fruitful when dealing with problems where one is not interested in recognition/identification only, but also into monitoring the behavior of consumers and/or machines, for instance. Therefore, by means of probability estimates, one can take decisions to work better in a number of scenarios. In this paper, we propose a probabilistic-based Optimum Path Forest (OPF) classifier to handle with binary classification problems, and we show it can be more accurate than na\"ive OPF in a number of datasets. In addition to being just more accurate or not, probabilistic OPF turns to be another useful tool to the scientific community.
\end{abstract}

\section{Introduction}
\label{introduction}

Pattern recognition techniques aim at learning \emph{decision functions} that somehow partition the feature space into clusters of samples that share some sort of behavior. Additionally, it is expected the learned function can generalize well over unseen data. Depending on the amount of information used concerning the learning process, decision functions (i.e. classifiers) are usually divided into three main categories: (i) \emph{supervised}, (ii) \emph{semi-supervised} and (iii) \emph{unsupervised}~\cite{Duda:00}. While the former approaches make use of a fully-labeled training set, semi-supervised approaches consider a partial-labeled data only. Finally, unsupervised techniques have no knowledge about training samples. Such techniques are commonly referred to clustering.

\begin{sloppypar}
Classification techniques are usually divided according to their output as well~\cite{AL-AniJAIR:02}: (i) \emph{abstract}, (ii) \emph{ranking} and (iii) \emph{confidence}. Abstract-based classifiers refer to the great majority of techniques, which output a label (usually an integer number) to each sample to be classified. Ranking-driven approaches may also output labels, but all possible outputs considered to a given sample are queued using some sort of heuristic, which are applied for different purposes. Finally, confidence-oriented techniques output some confidence value that is related to the \emph{probability} of some sample to be assigned to a given label. This last category concerns with the so-called \emph{probabilistic classifiers}.
\end{sloppypar}

Probabilistic techniques play an important role in machine learning, since they extend the classification process to a greater range than simply labels. Very often we face problems where it is desirable to obtain some probability than just the label itself. Consider the problem of theft identification in energy distribution systems. Electrical power companies consider much more fruitful to monitor the probability of a certain user to become a thiefer along the time instead of purely identifying such user. With the probabilities over time in hands, the company can take some preventive approach, which can be much more cost-effective than just punishing the user.

Fortunately, we have a considerable number of probabilistic-driven techniques in the literature. A seminal work conducted by Platt~\cite{PlattALMC:99} extended the well-known Support Vector Machines (SVMs), which were first designed to handle abstract outputs, to probabilistic classification. The idea is quite simple: to use SVMs' outputs (labels) to feed a logistic function. Therefore, the initial outputs are mapped within the range $[0,1]$. However, in order to cope with problems related to different quantities (SVMs' outputs before taking the signal to consider the final label), the author considered to use an optimization process over the whole training set in order to find out variables that regularize the label-probability mapping process. This technique is often referred to as ``Platt Scaling".

Later on, Niculescu-Mizil and Caruana~\cite{Niculescu-MizilICML:05} presented a very interesting comparison between Platt Scaling and Isotonic Regression to obtain probabilistic outputs concerning SVMs. Their work was motivated by the fact logistic functions may work well for several situations, but it may not be appropriate to others. Roughly speaking, Isotonic Regression aims at learning a function that is constrained to be monotonically increasing (isotonic), and it its fed with SVMs' real-valued outputs (i.e. before taking the signal of the function to classify a sample as positive or negative). The authors concluded Platt Scaling works better with small-sized datasets, and since Isotonic Regression is more prone to overfitting, it is recommended to be applied over large datasets.

Zadrozny and Elkan~\cite{ZadroznyICML:01} proposed to obtain probability estimates considering Decision Trees (DTs) and na\"ive Bayesian classifiers. The authors adopted smoother probability estimates for DTs, i.e., they adjust them to be less extreme. Smoothing is an interesting tool when dealing with probability estimation, since some methods may push probabilities away from the range $[0,1]$, and others adjust probabilities to be closer to $0.5$ (e.g. Laplace correction), which may not be interesting when classes are not equiprobable (in practice, they are not in real-world scenarios). Soon after, the very same group of authors extended their work to handle multiclass-oriented problems~\cite{ZadroznyKDD:02}. Other recent works can be referred as well~\cite{NapoliICAISC:15,SoundararajanCGF:15,SchleifESANN:15}, but they mainly focus on the application of probabilistic classifiers or comparison studies only, not on new theories or approaches.

Some years ago, a group of authors introduced the Optimum-Path Forest classifier (OPF), which is a framework to the design of graph-based classifiers that comprises supervised~\cite{PapaISVC:08,PapaIJIST:09,PapaPR:12} , semi-supervised~\cite{AmorimSIBGRAPI:14,AmorimPR:16} and unsupervised versions~\cite{RochaIJIST:09}. Roughly speaking, an OPF classifier models the problem of pattern recognition as a graph partition task, where some key samples (\emph{prototypes}) compete among themselves in order to conquer the remaining samples by means of a reward-compensation process. At the final, we have an optimum-path forest, which is essentially a collection of optimum-path trees (clusters) rooted at each prototype sample. OPF has demonstrated very suitable results in a number of applications, being usually faster than SVMs for training, tough with similar or even better accuracy.

However, na\"ive OPF works with abstract outputs only. Also, as far as we know, there is only one very recent work that considered confidence-based OPF, but not for probability estimates~\cite{FernandesCIARP:15}. That work proposed to learn the confidence level (reliability) of each training sample when classifying others. Additionally, the cost-function used for conquering purposes was adapted to consider such reliability level. The authors showed the proposed confidence-based OPF works better in datasets with high concentration of overlapped samples. Furthermore, to the best of our knowledge, there is no probabilistic-driven OPF to date, which turns to be the main contribution of this work, i.e. to fill the lack of research regarding confidence-based outputs with respect to OPF classifiers. The proposed approach, initially designed to cope with binary-oriented classification problems, is compared against na\"ive OPF in different scenarios, showing very suitable results. The remainder of this paper is organized as follows. Sections~\ref{s.opf} and~\ref{s.prob_opf} present the OPF theoretical background and the probabilistic-driven approach, respectively. Section~\ref{s.methodology} discusses the methodology and Section~\ref{s.experiments} presents experiments. Finally, Section~\ref{s.conclusions} states conclusions and future works.

\section{Optimum-Path Forest}
\label{s.opf}

Let ${\cal D}={\cal D}^{tr}\cup{\cal D}^{ts}$ be a $\lambda$-labeled dataset such that ${\cal D}^{tr}$ and ${\cal D}^{ts}$ stand for the training and testing sets, respectively. Additionally, let $\textbf{s}\in{\cal D}$ be an $n$-dimensional sample that encodes features extracted from a certain data, and $d(\textbf{s},\textbf{v})$ be a function that computes the distance between two samples $\textbf{s}$ e $\textbf{v}$, $\textbf{v}\in{\cal D}$.

Let ${\cal G}^{tr}=({\cal D}^{tr},{\cal A})$ be a graph derived from the training set, such that each node $\textbf{v}\in{\cal D}^{tr}$ is connected to every other node in ${\cal D}^{tr}\backslash\{\textbf{v}\}$, i.e. ${\cal A}$ defines an adjacency relation known as \emph{complete graph}, in which the arcs are weighted by function $d(\cdot,\cdot)$. We can also define a path $\pi_s$ as a sequence of adjacent and distinct nodes in ${\cal G}^{tr}$ with terminus at node $\textbf{s}\in{\cal D}^{tr}$. Notice a \emph{trivial path} is denoted by $\langle s\rangle$, i.e. a single-node path.

Let $f(\pi_s)$ be a path-cost function that essentially assigns a real and positive value to a given path $\pi_s$, and ${\cal S}$ be a set of prototype nodes. Roughly speaking, OPF aims at solving the following optimization problem:

\begin{equation}
	\label{e.opf_optimization}
	\min f(\pi_s),\ \forall\ \textbf{s}\in{\cal D}^{tr}.
\end{equation}
The good point is that one does not need to deal with mathematical constraints, and the only rule to solve Equation~\ref{e.opf_optimization} concerns that all paths must be rooted at ${\cal S}$. Therefore, we must choose two principles now: how to compute ${\cal S}$ (prototype estimation heuristic) and $f(\pi)$ (path-cost function).

Since prototypes play a major role, Papa et al.~\cite{PapaIJIST:09} proposed to position them at the regions with the highest probabilities of misclassification, i.e. at the boundaries among samples from different classes. In fact, we are looking for the nearest samples from different classes, which can be computed by means of a Minimum Spanning Tree (MST) over ${\cal G}^{tr}$. The MST has interesting properties, which ensure OPF can be errorless during training when all arc-weights are different to each other~\cite{AlleneIVC:10}.

Finally, with respect to the path-cost function, OPF requires $f$ to be a smooth one~\cite{FalcaoIEEEPAMI:04}. Previous experience in image segmentation led the authors to use a chain code-invariant path-cost function, that basically computes the maximum arc-weight along a path, being denoted as $f_{max}$ and given by:

\begin{eqnarray}
	\label{e.fmax}
	f_{max}(\langle s\rangle) & = & \left\{ \begin{array}{ll}
	0 & \mbox{if $\textbf{s}\in {\cal S}$} \\
	+\infty & \mbox{otherwise,}
	\end{array}\right. \nonumber \\
	f_{max}(\pi_s \cdot (\textbf{s},\textbf{t})) & = & \max\{f_{max}(\pi_s),d(\textbf{s},\textbf{t})\}, 
\end{eqnarray}
where $\pi_s \cdot (\textbf{s},\textbf{t})$ stands for the concatenation between path $\pi_s$ and arc $(\textbf{s},\textbf{t})\in{\cal A}$. In short, by computing Equation~\ref{e.fmax} for every sample $\textbf{s}\in{\cal D}^{tr}$, we obtain a collection of optimum-path trees (OPTs) rooted at ${\cal S}$, which then originate an optimum-path forest. A sample that belongs to a given OPT means it is more strongly connected to it than to any other in ${\cal G}^{tr}$. Roughly speaking, the OPF training step aims at solving Equation~\ref{e.fmax} in order to build the optimum-path forest.

The next step concerns the testing phase, where each sample $\textbf{t}\in{\cal D}^{ts}$ is classified individually as follows: $\textbf{t}$ is connected to all training nodes from the optimum-path forest learned in the training phase, and it is evaluated the node $\textbf{v}^\ast\in{\cal D}^{tr}$ that conquers $\textbf{t}$, i.e. the one that satisfies the following equation: 

\begin{equation}
	\label{e.opfcpl_classification}
	C_\textbf{t}=\argminA_{\textbf{v}\in{\cal D}^{tr}}\max\{C_\textbf{v},d(\textbf{v},\textbf{t})\}.
\end{equation}
The classification step simply assigns $L(\textbf{t})=\lambda(\textbf{v}^\ast)$. Roughly speaking, the testing step aims at finding the training node $\textbf{v}$ that minimizes $C_\textbf{t}$.

It is worth noting that OPF is not a distance-based classifier, but instead it uses the ``power of connectivity"\ among samples. The OPF with complete graph degenerates to a nearest neighbor classifier only when all training samples are prototypes. Actually, such situation is considerably difficult to face, thus indicating a high degree of overlapping among samples, which means the features used for that specific problem may not be adequate enough to describe it.

\section{Probabilistic Optimum-Path Forest}
\label{s.prob_opf}

The probabilistic OPF is inspired in the Platt Scaling approach, which basically ends up mapping the SVMs' output to probability estimates. Therefore, before introducing the proposed approach, one must master the Platt Scaling mechanism.

Considering the labeled dataset ${\cal D}$ described in Section~\ref{s.opf}, let us assume each sample $\textbf{x}_i\in{\cal D}$ can be assigned to a class label $y_i\in\{-1,+1\}$, $i=1,2,\ldots,\left|{\cal D}\right|$. Platt proposed to approximate the posterior class probability $P(y_i=1|\textbf{x}_i)$ as follows~\cite{PlattALMC:99}:

\begin{equation}
\label{e.probability}
	P(y_i=1|\textbf{x}_i)\approx P_{A,B}(f_i)\equiv\frac{1}{1+\exp{(Af_i+B)}},
\end{equation}
where $f_i$ stands for the output (decision function) of SVMs concerning sample $\textbf{x}_i$. Let $\theta=(A^\ast,B^\ast)$ be the best set of parameters that can be determined by the following maximum likelihood problem:

\begin{equation}
\label{e.cost_function}
	\argminA_{\theta}F(\theta)=-\sum_{i=1}^m(y_i\log(p_i)+(1-y_i)\log(1-p_i)),
\end{equation}
where $p_i=P_{A,B}(f_i)$ and $m$ denotes the number of samples to be considered. Essentially, the above equation stands for the cost function of the well-known Logistic Regression classifier.

In order to avoid overfitting, Platt proposed to regularize Equation~\ref{e.cost_function} as follows:

\begin{equation}
\label{e.cost_function_regularized}
	\argminA_{\theta}F(\theta)=-\sum_{i=1}^m(t_i\log(p_i)+(1-t_i)\log(1-p_i)),
\end{equation}
where $t_i$ is formulated as follows:

\begin{eqnarray}
	\label{e.ti}
t_i & = & \left\{ \begin{array}{ll}
	\frac{N_++1}{N_++2} & \mbox{if $y_i=+1$} \\
	\frac{1}{N_-+2} & \mbox{if $y_i=-1$.}
	\end{array}\right.
\end{eqnarray}
In the above formulation, $N_+$ and $N_-$ stand for the number of positive and negative samples, respectively. In short, $t_i$ can be used to handle unbalanced datasets as well.

Since the cost assigned to each sample during training and classification with OPF is positive (Equation~\ref{e.opfcpl_classification}), we need some minor adjustments with respect to Equation~\ref{e.probability}, which can be rewritten to accommodate OPF requirements:

\begin{equation}
\label{e.probability_opf}
	P(y_i=1|\textbf{x}_i)\approx P_{A,B}(C_i)\equiv\frac{1}{1+\exp{(Ay_iC_i+B)}},
\end{equation}
where $C_i$ stands for the cost assigned to sample $\textbf{x}_i$ during OPF training or classification step. Basically, we ended up replacing $f_i$ by $y_iC_i$, since the cost function $C_i$ is not signed, while $\sgn(f_i)\in\{-,+\}$. 

The rationale behind the proposed approach is to assume the lower the cost assigned to sample $\textbf{x}_i$, i.e. $C_i$, the higher the probability of that sample be correctly classified. A similar idea is used by Platt, since the greater $f_i$ (i.e. the farthest a sample is from the decision boundary), the more likely that sample belongs to class $+1$ (positive side) or $-1$ (negative side). In addition, probabilistic OPF also makes use of Equation~\ref{e.cost_function_regularized}, but now with $p_i=P_{A,B}(C_i)$.

Almost a decade later the seminal work of Platt, Lin et al.~\cite{LinML:07} highlighted some numerical instabilities related to Equation~\ref{e.cost_function_regularized}:

\begin{itemize}
	\item we know that $\log$ and $\exp$ functions can easily  cause an overflow, since $\exp(Af_i+B)\rightarrow\infty$ when $Af_i+B$ is large enough. Additionally, $\log(p_i)\rightarrow-\infty$ when $p_i\rightarrow 0$.
	\item according to Goldberg~\cite{GoldbergACS:91}, $1-p_i=1-\frac{1}{1+\exp(Af_i+B)}$ is a ``catastrophic cancellation"\ when $p_i$ is close to one. Such term arises from the fact we need to subtract two relatively close number that are already results of previous floating-point operations. Lin et al.~\cite{LinML:07} described an interesting example: suppose $f_i=1$ and $(A,B)=(-64,0)$. In this case, $1-p_i$ returns 0, but its equivalent formulation $\frac{\exp(Af_i+b)}{1+\exp(Af_i+B)}$ gives a more accurate result. Also, the very same group of authors stated the aforementioned catastrophic cancellation induces most of the $\log(0)$ occurrences.
\end{itemize}

In order to deal with the aforementioned situation, Lin et al.~\cite{LinML:07} proposed to reformulate the cost function $F(\theta)$ as follows:

\begin{eqnarray}
	\label{e.ti_modified1}
F(\theta) & = & -\sum_{i=1}^m(t_i\log(p_i)+(1-t_i)\log(1-p_i))\\
	\label{e.ti_modified2}
& = & -\sum_{i=1}^m((t_i-1)(q_i)+\log(1+\exp(q_i)))\\
	\label{e.ti_modified3}
& = & -\sum_{i=1}^m(t_iq_i+\log(1+\exp(-Af_i-B))),
\end{eqnarray}
where $q_i=Af_i+B$. Therefore, considering the above formulation, $1-p_i$ and $\log(0)$ do not happen\footnote{Please, consider taking a look at the work of Lin et al.~\cite{LinML:07} for a more detailed explanation about the mathematical formulation.}.

However, even if using Equations~\ref{e.ti_modified2} and~\ref{e.ti_modified3}, the overflow problem may still occur. In order to cope with such problem, Lin et al.~\cite{LinML:07} proposed to apply Equation~\ref{e.ti_modified3} when $Af_i+B\geq 0$; otherwise, one should use Equation~\ref{e.ti_modified2}. Similarly, we adopted the very same procedure concerning probabilistic OPF, hereinafter called P-OPF. In short, one can implement P-OPF by just changing $f_i$ by $y_iC_i$ in Equations~\ref{e.ti_modified2} and~\ref{e.ti_modified3}, $i=1,2,\ldots,m$.

After learning parameters $A$ and $B$, we then compute the probability of each sample to belong to class $+1$, i.e. $P(y_i=1|\textbf{x}_i)$. If $P(y_i=1|\textbf{x}_i)\geq \Theta$, then P-OPF assigns the label $+1$ to that sample; otherwise the sample is assigned to class $-1$. In this work, we adopted $\Theta=0.5$, since it models a single chance. However, one can easily fine-tune that threshold using a linear-search or any other optimization algorithm.

\section{Methodology}
\label{s.methodology}

In this section, we present the methodology used to compare P-OPF against nai\"ive OPF. Although we could consider any other probabilistic classifier for comparison purposes, the main idea of this work does not concern with outperforming other techniques, but to propose a probabilistic OPF technique instead.

In order to fine-tune parameters $A$ and $B$, we employed four different optimization methods, being three of them based on meta-heuristics, and another one purely mathematical. In regard to the meta-heuristic-driven techniques, we opted to use Bat Algorithm (BA)~\cite{YangBA:12}, Firefly Algorithm (FFA)~\cite{YangFFA:10} and Particle Swarm Optimization (PSO)~\cite{Kennedy:01}, and with respect to the another mathematical method we used the Nelder-Mead (NM)~\cite{NelderTCJ:65}. The main reason to use the aforementioned techniques concerns their very good effectiveness in a number of problems in the literature.

In order to study the behavior of P-OPF under different scenarios, we used three synthetic datasets (synhetic0, synhetic2, synhetic3), two datasets concerning energy theft detection (comercial and industrial)\cite{Pereira:2016ETD}, as well as nine public benchmarking datasets\footnote{\url{http://www.csie.ntu.edu.tw/~cjlin/libsvmtools/datasets/}}. These datasets have been frequently used in the evaluation of different classification methods. Table~\ref{tab.datasets} presents the main characteristics of each dataset. 

\begin{table}[hb]
\small
\begin{center}
\begin{tabular}{l c c c}
\hline\hline
{\bf Dataset} & {\bf \# samples} & {\bf \# features} & {\bf \# classes}\\
\hline
australian       &      $690$           &        $14$           &    $2$\\
comercial        &      $4,952$          &        $8$            &    $2$\\
industrial       &      $3,182$          &        $8$            &    $2$\\
breast           &	$683$	        & 	 $10$           &    $2$\\ 
colon\_cancer    &      $62$            &        $2,000$        &    $2$\\
diabetes         &      $768$           &        $8$            &    $2$\\ 
fourclass        &      $862$           &        $2$            &    $2$\\
heart            &      $270$           &        $13$           &    $2$\\
ionosphere       &      $351$           &        $34$           &    $2$\\
ionosphere\_scale&      $351$           &        $34$           &    $2$\\
liver            &      $345$           &        $6$            &    $2$\\
synthetic0       &      $500$           &        $2$            &    $2$\\
synthetic2       &      $1,000$         &        $2$            &    $2$\\
synthetic3       &      $200$           &        $2$            &    $2$\\
\hline
\end{tabular}
\end{center}
\caption{\label{tab.datasets} Information about the benchmarking datasets used in this work.}
\end{table}

In addition, we randomly divided each dataset into two disjoint sets: training ($Z_1$) and testing ($Z_2$). The training and testing set sizes were defined as $25\%$ and $75\%$, respectively\footnote{Notice these percentages were empirically chosen.}. The experimental setup was conducted using a cross-validation procedure with $20$ runnings. In order to compare P-OPF and OPF, we computed the mean accuracy and execution time for the further usage of the Wilcoxon signed rank test~\cite{Wilcoxon:45} with significance of $0.05$. 

In regard to the optimization techniques, we used $20$ agents (initial solutions) concerning BA, FFA and PSO, as well as $400$ iterations for convergence. The search space for  $A\times B$ was defined within $[-10,10]\times[-10,10]$. Table~\ref{t.parameters} presents the parameter setup concerning the meta-heuristic techniques. Once again, these values have been empirically chosen.

\begin{table}[h]
		    \begin{center}
		    \begin{tabular}{  c | c  }
		      \hline \hline
		       \textbf{Technique} & \textbf{Parameters}\\ \hline 
		        BA & $q_{min}=0.0$, $q_{max}=1.0$, $\alpha=\gamma=1.0$ \\
			FFA & $\gamma=1.0$, $\beta=0.9$, $\alpha=0.7$ \\
		        PSO & $c_1=c_2=2.0$, $w=0.5$\\ 
			NM  & $p = 0.001$, $max_{it} = 1000$ \\  \hline 
		    \end{tabular}		    
		    \newline  
		    \caption{\label{t.parameters} Parameter configuration regarding meta-heuristic techniques.}		    		    
		    \end{center}		    
		   \end{table}

\begin{figure}[!htb]
  \begin{center}
    \includegraphics[scale=0.3]{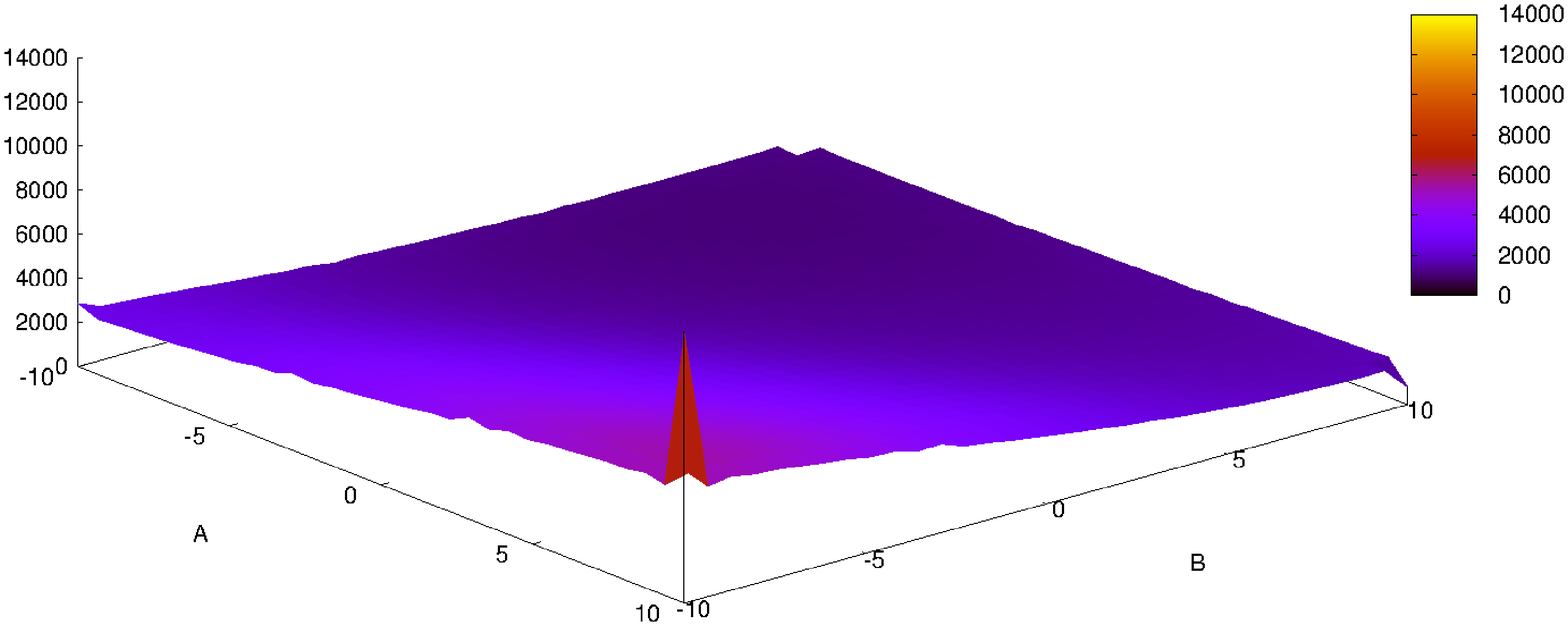}\\
    (a)\\\vspace{0.5cm}
    \includegraphics[scale=0.3]{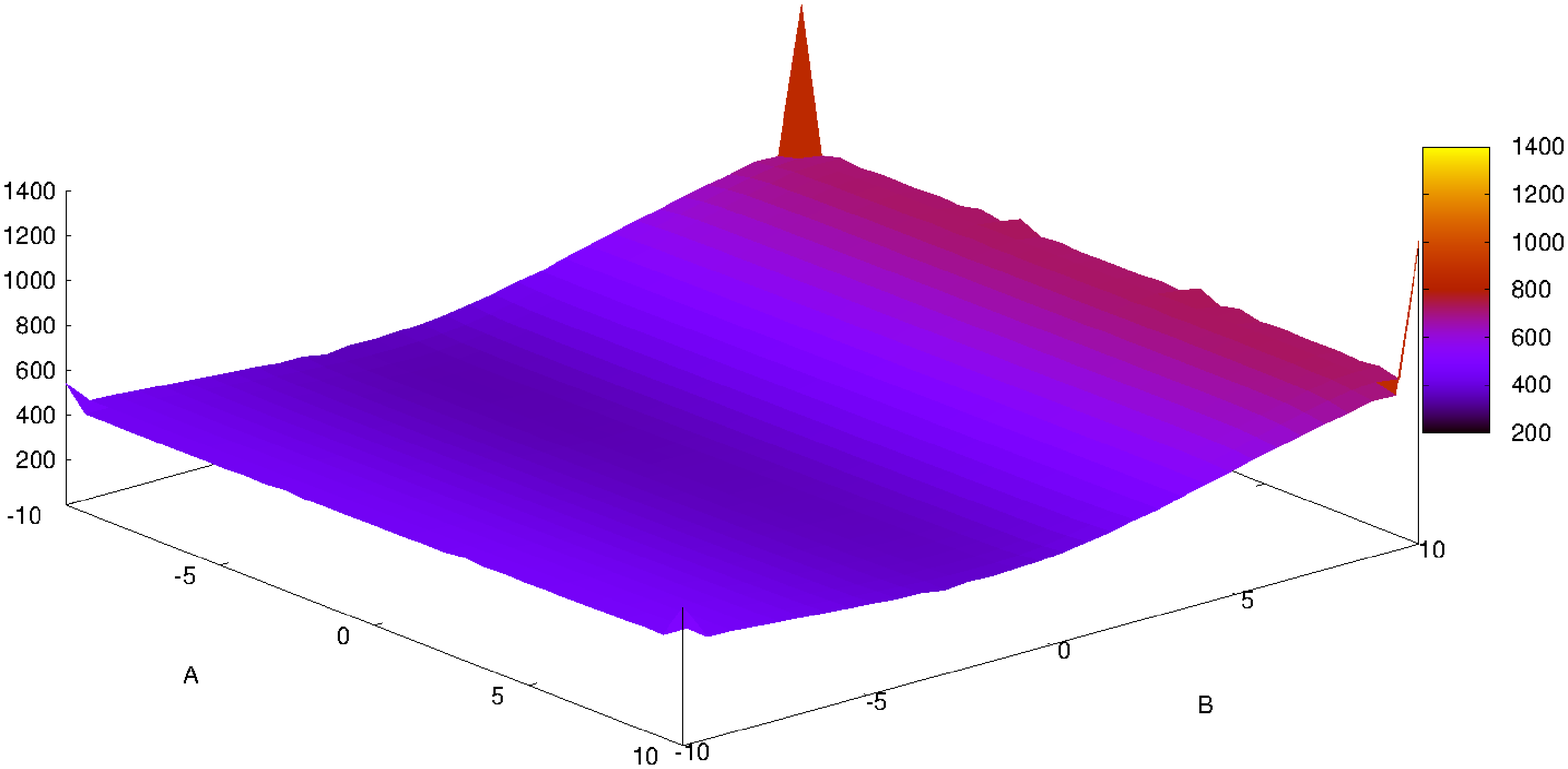}\\
    (b)
  \end{center} 
  \caption{Fitness landscape functions concerning: (a) industrial and (b) diabetes datasets.}
  \label{f.graph}
\end{figure}

In order to justify the application of a conventional optimization method, we plotted the fitness landscape of the optimization function $F(\theta)$ built under a grid-search over the search space $A\times B$. Figure~\ref{f.graph} depicts the fitness landscape concerning industrial and diabetes datasets. Due to the smoothness and apparently quasi-convexity, we opted to employ a conventional technique for optimization purposes (i.e. Nelder-Mead). 

\section{Experiments and Results}
\label{s.experiments}

In this section, we present the experimental results regarding the probabilistic OPF. Tables~\ref{t.acc} and~\ref{t.time} present the mean accuracy and computational load (seconds) concerning the compared methods\footnote{We employed an accuracy measure proposed by Papa et al.~\cite{PapaIJIST:09} that considers unbalanced datasets}. The most accurate techniques considering the Wilcoxon test are highlighted in bold. 

\begin{table}[!htb]
\centering

\caption{Mean accuracy considering na\"ive OPF, P-OPF and its variations under different optimization techniques.}
\label{t.acc}
\resizebox{\textwidth}{!}{
\begin{tabular}{l|lllll}
& \multicolumn{5}{c}{Accuracy(\%)}                  \\
\hline\hline
Dataset & OPF & P-OPF-BA & P-OPF-FFA & P-OPF-PSO & P-OPF-NM \\
\hline
australian& $46.96\pm6.55$& $\mathbf{52.50\pm6.84}$& $\mathbf{53.31\pm6.45}$& $\mathbf{52.43\pm6.87}$& $\mathbf{53.31\pm6.45}$ \\ 
comercial& $\mathbf{87.66\pm3.04}$ & $78.00\pm25.44$ & $43.34\pm34.45$ & $\mathbf{87.66\pm3.04}$ & $\mathbf{87.66\pm3.04}$\\
industrial& $\mathbf{97.03\pm0.55}$& $\mathbf{97.03\pm0.55}$& $36.69\pm36.91$& $\mathbf{97.02\pm0.54}$& $\mathbf{97.02\pm0.54}$ \\ 
breast\_cancer& $\mathbf{95.83\pm0.80}$& $86.73\pm20.32$& $75.81\pm36.32$& $\mathbf{95.83\pm0.80}$& $\mathbf{95.83\pm0.80}$ \\
colon\_cancer& $\mathbf{61.70\pm5.49}$& $50.00\pm14.58$& $52.77\pm14.28$& $50.00\pm14.58$& $\mathbf{61.06\pm6.75}$ \\ 
diabetes& $\mathbf{61.39\pm11.10}$& $55.47\pm13.05$& $51.63\pm14.39$& $43.18\pm12.53$& $43.18\pm12.53$ \\ 
fourclass& $\mathbf{50.43\pm11.89}$& $47.67\pm11.76$& $49.18\pm11.98$& $\mathbf{50.05\pm12.02}$& $\mathbf{50.05\pm12.02}$ \\ 
heart& $\mathbf{65.32\pm4.98}$& $50.44\pm7.58$& $48.13\pm7.08$& $54.48\pm5.92$& $54.48\pm5.92$ \\ 
ionosphere& $\mathbf{85.64\pm3.25}$& $55.57\pm13.68$& $43.33\pm10.32$& $81.29\pm3.17$& $83.30\pm3.11$ \\ 
ionosphere\_scale& $\mathbf{85.04\pm3.12}$& $60.98\pm12.39$& $53.26\pm14.59$& $79.62\pm8.95$& $80.61\pm8.90$ \\ 
liver& $\mathbf{61.00\pm2.36}$& $54.02\pm9.62$& $45.02\pm10.52$& $60.50\pm2.31$& $\mathbf{60.50\pm2.31}$ \\ 
synthetic0& $48.72\pm3.67$& $50.77\pm3.66$& $50.29\pm3.74$& $\mathbf{51.57\pm3.37}$& $\mathbf{51.57\pm3.37}$ \\ 
synthetic2& $\mathbf{51.99\pm5.55}$& $49.84\pm6.27$& $46.77\pm5.27$& $49.92\pm6.27$& $49.92\pm6.27$ \\ 
synthetic3& $42.78\pm5.79$& $50.86\pm9.81$& $44.83\pm8.36$& $\mathbf{58.34\pm4.47}$& $55.23\pm8.18$ \\
\end{tabular}
}
\end{table}

The proposed P-OPF obtained the best results for nine datasets, while na\"ive OPF achieved the best result for eleven datasets. In three out nine datasets, P-OPF obtained the top results. The results are quite interesting, since P-OPF was able to improve OPF for datasets, besides being able to output probability estimates. Considering some other datasets, although P-OPF did not outperform OPF, the former achieved considerably close results, which is somehow interesting, since P-OPF can obtain similar accuracies compared to OPF, but being able to output probabilities as well.

In regard to the optimization techniques, NM obtained the best results for eight datasets, closely followed by PSO, which obtained the best results in seven datasets. However, if we consider a trade-off between computational load and accuracy, NM has been the best optimization approach, since it has a lower computational cost. The good performance of NM is mainly due to the smoothness of the objective functions. Table~\ref{t.time} presents the mean computational load in seconds concerning na\"ive OPF and P-OPF with parameters fine-tuned with the optimization techniques. Since BA, FFA and PSO are swarm-based techniques, which means they update all possible solutions (agents) at each iteration, they are much more costly than NM.

\begin{table*}[h]
\centering
\small
\caption{Computational load concerning na\"ive OPF, P-OPF and its variations under different optimization techniques.}
\label{t.time}
\begin{tabular}{l|lllll}
& \multicolumn{5}{c}{Time(s)}                  \\
\hline\hline
Dataset & OPF & P-OPF-BA & P-OPF-FFA & P-OPF-PSO & P-OPF-NM \\
\hline
australian& $0.00\pm0.00$ & $0.17\pm0.02$& $0.20\pm0.01$& $0.10\pm0.01$ & $0.02\pm0.00$\\ 
industrial& $0.06\pm0.00$ & $1.00\pm0.15$& $0.98\pm0.14$& $0.97\pm0.09$ & $0.30\pm0.00$ \\ 
industrial& $0.05\pm0.00$ & $0.96\pm0.12$& $0.91\pm0.11$& $0.88\pm0.11$ & $0.26\pm0.06$ \\ 
breast\_cancer& $0.01\pm0.00$& $0.23\pm0.05$& $0.30\pm0.06$& $0.32\pm0.01$ & $0.12\pm0.01$\\ 
colon\_cancer& $0.00\pm0.00$ & $0.10\pm0.01$& $0.10\pm0.01$& $0.10\pm0.01$ & $0.01\pm0.00$\\ 
diabetes& $0.01\pm0.00$& $0.25\pm0.03$& $0.32\pm0.05$& $0.27\pm0.01$ & $0.04\pm0.00$\\ 
fourclass&$0.01\pm0.00$ &  $0.27\pm0.03$& $0.28\pm0.03$& $0.28\pm0.03$ & $0.05\pm0.00$\\ 
heart& $0.00\pm0.00$& $0.17\pm0.03$& $0.18\pm0.02$& $0.18\pm0.01$ &  $0.02\pm0.00$\\ 
ionosphere& $0.00\pm0.00$ & $0.21\pm0.04$& $0.21\pm0.04$& $0.20\pm0.02$ & $0.03\pm0.00$\\ 
ionosphere\_scale& $0.00\pm0.00$& $0.18\pm0.02$& $0.19\pm0.03$& $0.2\pm0.02$ & $0.01\pm0.00$\\ 
liver& $0.00\pm0.00$ & $0.17\pm0.01$& $0.18\pm0.01$& $0.16\pm0.01$ & $0.01\pm0.00$\\ 
synthetic0& $0.00\pm0.00$& $0.19\pm0.01$& $0.19\pm0.01$& $0.20\pm0.01$ & $0.02\pm0.00$\\ 
synthetic2& $0.02\pm0.00$ & $0.31\pm0.02$& $0.30\pm0.02$& $0.29\pm0.03$ & $0.10\pm0.03$\\ 
synthetic3& $0.00\pm0.00$& $0.12\pm0.01$& $0.13\pm0.01$& $0.11\pm0.01$ & $0.01\pm0.00$\\ 
\end{tabular}
\end{table*}

\begin{figure}[!htb]
\centerline{
  \begin{tabular}{cc}
  	  \includegraphics[width=6.5cm,height=3.5cm]{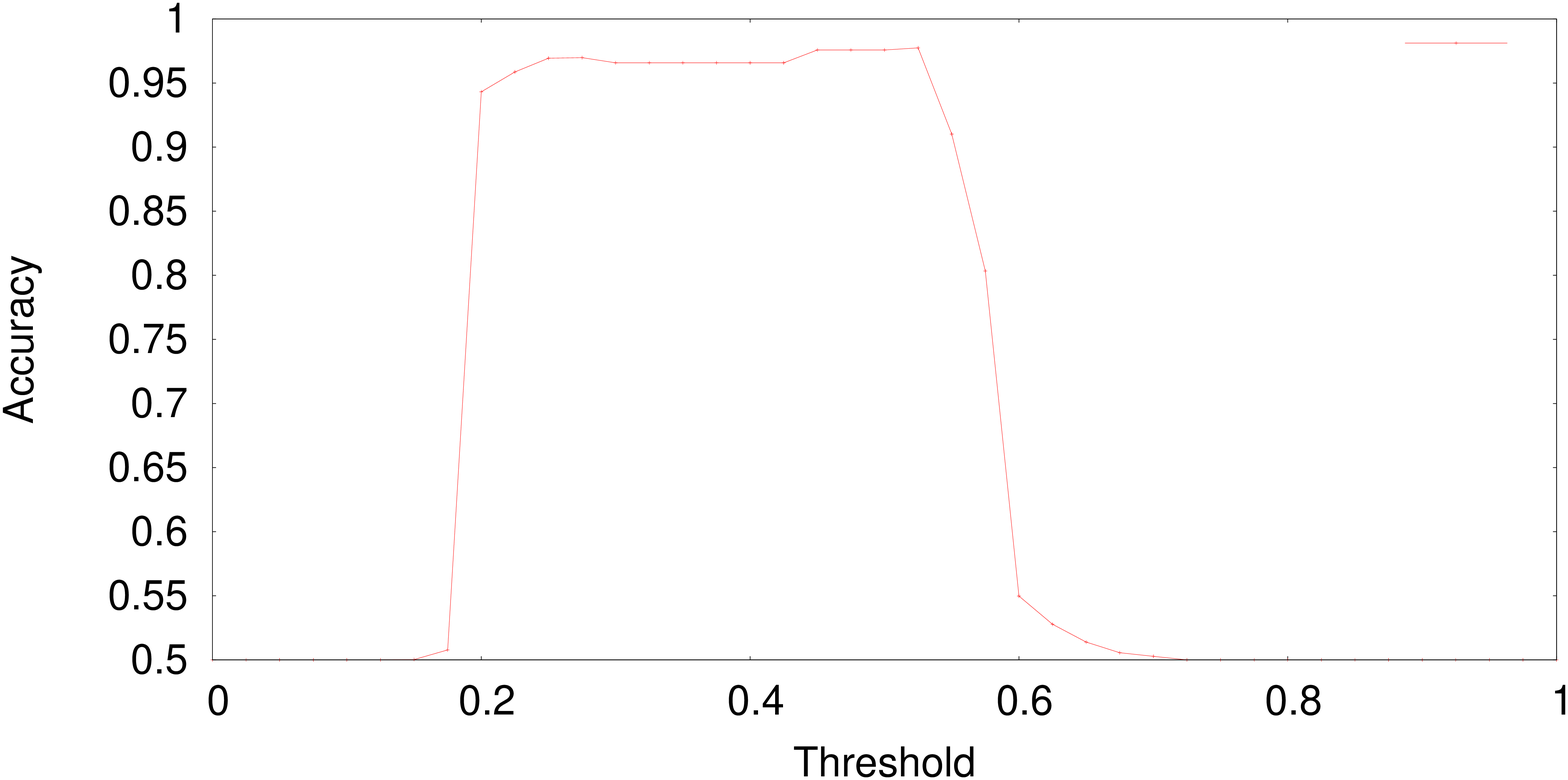} \hspace{-0.6cm} &
    \includegraphics[width=6.5cm,height=3.5cm]{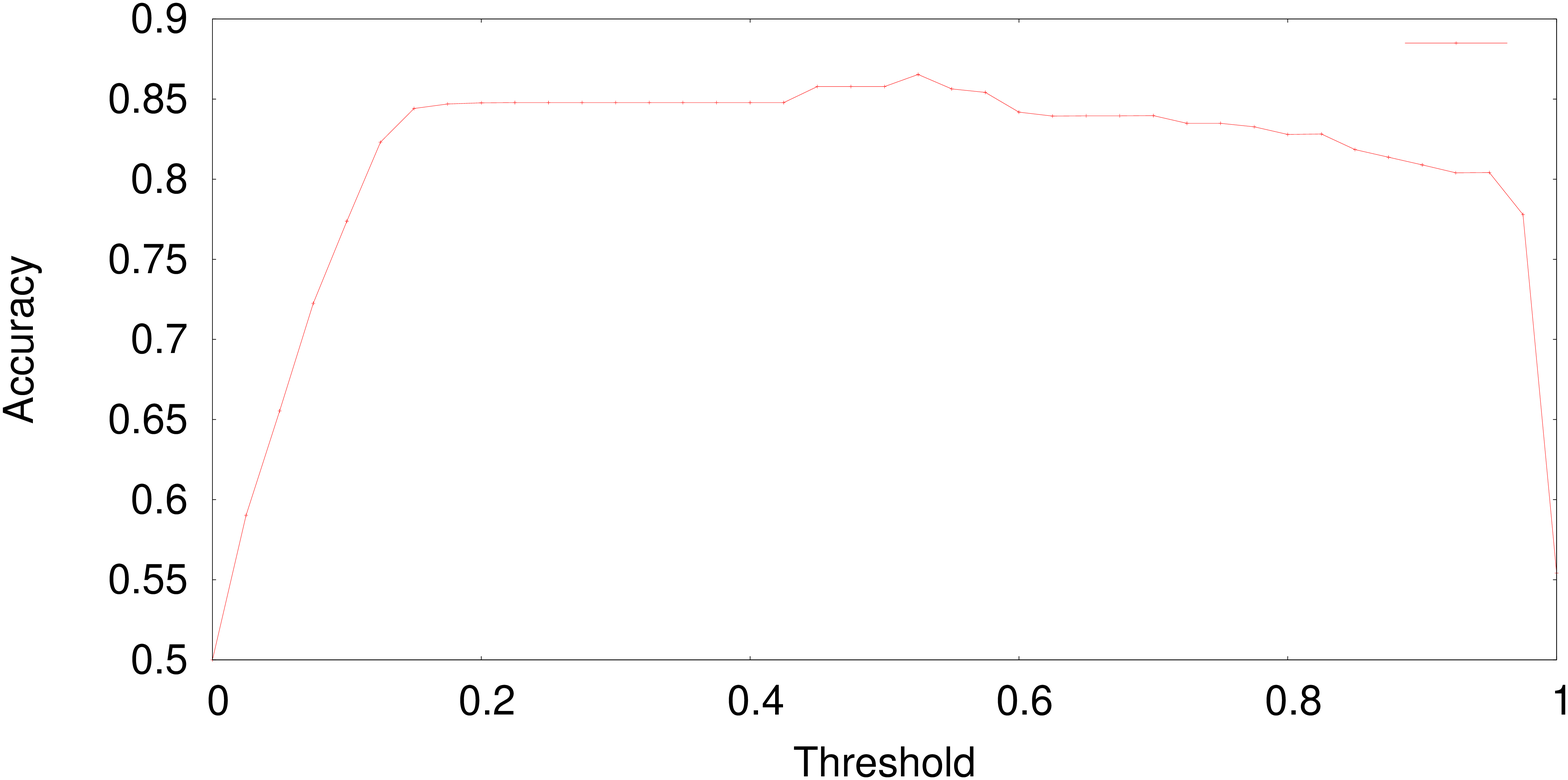}\\
    \hspace{1cm}(a) & \hspace{0.5cm} (b)\\
  \end{tabular}
}
  \caption{Influence of the threshold parameter over: (a) breast\_cancer and (b) comercial datasets.}
  \label{f.threshold1}
\end{figure}

\begin{figure}[!htb]
\centerline{
  \begin{tabular}{cc}
    \includegraphics[width=6.5cm,height=3.5cm]{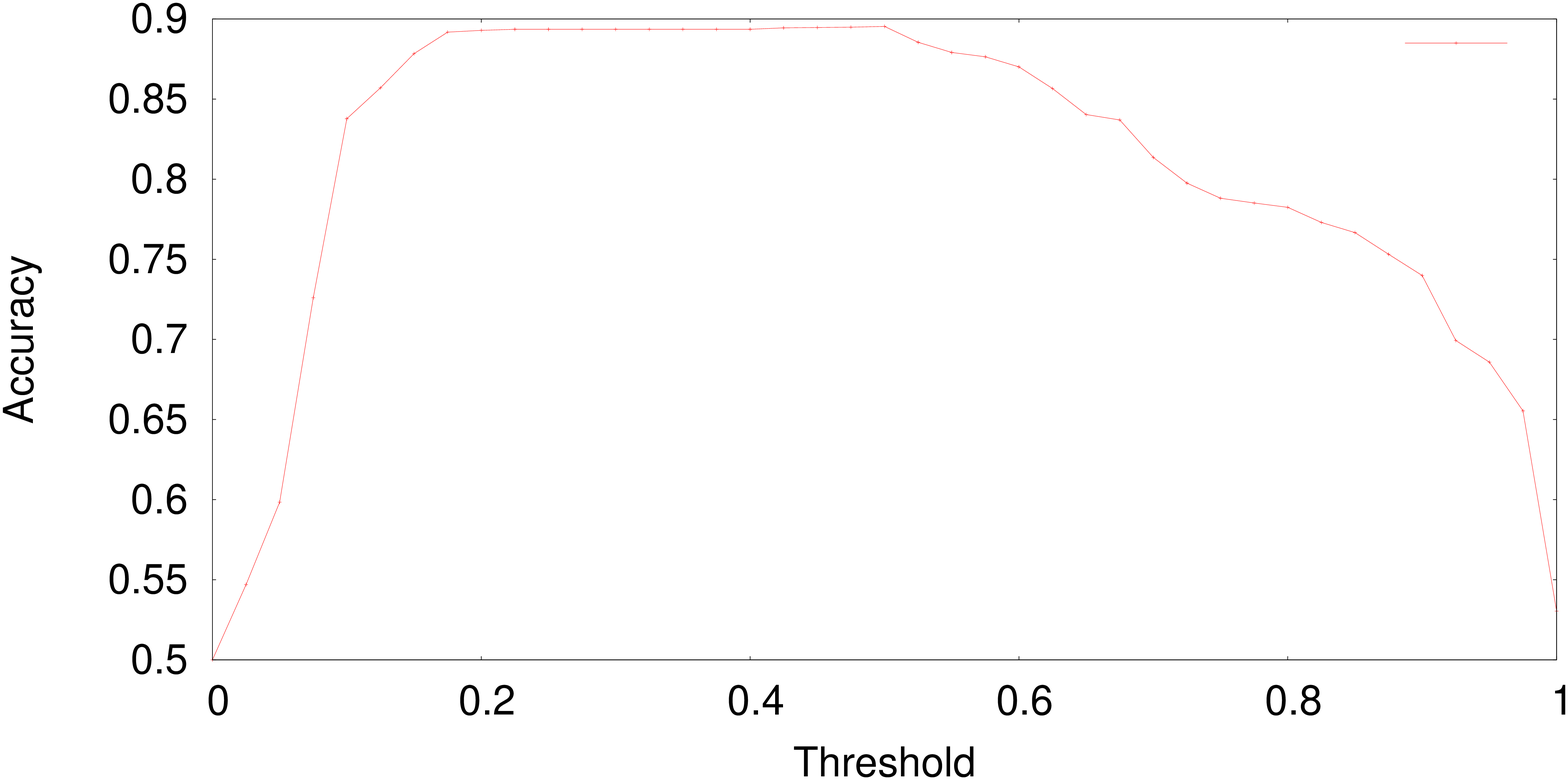} \hspace{-0.6cm} &
    \includegraphics[width=6.5cm,height=3.5cm]{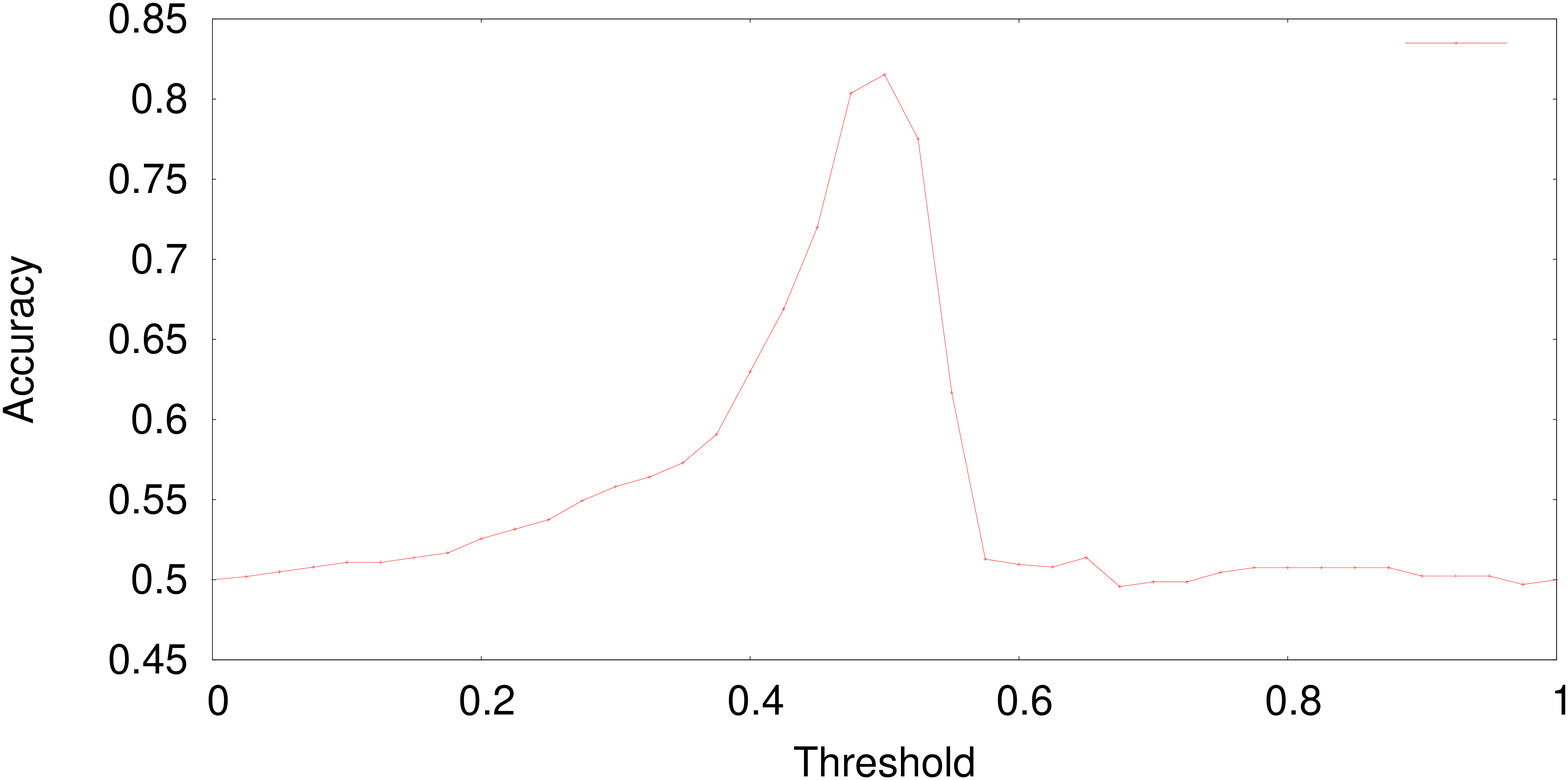}\\
	\hspace{1cm} (a) & \hspace{0.5cm} (b)\\
 \end{tabular}
}
  \caption{Influence of the threshold parameter over: (a) industrial and (b) ionosphere datasets.}
  \label{f.threshold2}
\end{figure}

Finally, we conducted an extra round of experiments to assess the influence of the threshold parameter $\Theta$. Figures~\ref{f.threshold1} and~\ref{f.threshold2} display the accuracy values over different thresholds considering four datasets: breast\_cancer, comercial, industrial and ionosphere. Clearly, one can observe that some datasets contain a certain plateau of accuracies considering different threshold values (Figures~\ref{f.threshold1}b and~\ref{f.threshold2}b), but for other datasets such plateau is smaller (Figure~\ref{f.threshold1}a) or even does not exist (Figure~\ref{f.threshold2}b). As aforementioned, such behaviour led us to use $\Theta=0.5$ for all datasets, since the same behaviour (or at least a similar one) has been observed for all datasets.

\section{Conclusions and Future Works}
\label{s.conclusions}

Probabilistic classification has been a topic of great interest concerning the machine learning community, mainly due to the lack of a more ``flexible"\ information rather than labels only. In this work, we cope with this problem by proposing a probabilistic OPF for binary classification problems, namely P-OPF. The results of the proposed P-OPF were compared against na\"ive OPF in a number of datasets, achieving suitable results in several of them. Also, we compared four optimization techniques to minimize a cost function aiming at learning its best parameters over the whole training set. 

In regard to future works, we aim at extending P-OPF for multi-class classification problems, as well as to consider other optimization techniques to fine-tune the new parameters that help minimizing the cost function. Also, we shall consider using the derivative of the cost function together with optimization techniques that require such computation explicitly.

\begin{acknowledgements}
The authors would like to thank Capes, CNPq grant \#306166/2014-3 and FAPESP grant \#2014/16250-9
\end{acknowledgements}

\bibliographystyle{spmpsci}      % mathematics and physical sciences
\bibliography{refs}   % name your BibTeX data base

\end{document}